\documentclass[10pt,twocolumn,letterpaper]{article}

\usepackage[pagenumbers,algorithms]{wacv}      

\usepackage{graphicx}
\usepackage{amsmath}
\usepackage{amssymb}
\usepackage{booktabs}

\usepackage[pagebackref,breaklinks,colorlinks]{hyperref}

\usepackage[capitalize]{cleveref}
\crefname{section}{Sec.}{Secs.}
\Crefname{section}{Section}{Sections}
\Crefname{table}{Table}{Tables}
\crefname{table}{Tab.}{Tabs.}

\usepackage{amssymb} 
\usepackage{xcolor}  
\usepackage{colortbl} 
\usepackage{graphicx} 
\usepackage{makecell} 
\usepackage{rotating} 

\definecolor{LightGrey}{rgb}{.9,.9,.9}
\definecolor{White}{rgb}{1.,0.,1.}
\definecolor{first}{rgb}{.8,.0,.0}
\definecolor{second}{rgb}{.0,.6,.0}
\definecolor{third}{rgb}{.0,.0,.8}
\definecolor{nbarrier}{RGB}{255, 120, 50}

\usepackage{amssymb}
\usepackage{xcolor}
\usepackage{graphicx}
\usepackage{booktabs}
\usepackage{amsmath}
\usepackage{amssymb}
\usepackage{multirow}
\usepackage{makecell}
\usepackage{array}
\newcommand{\PreserveBackslash}[1]{\let\temp=\\#1\let\\=\temp}
\newcolumntype{M}[1]{>{\PreserveBackslash\centering\rule{0pt}{20pt}}m{#1}} 

\usepackage[accsupp]{axessibility}  

\newlength\savewidth\newcommand\shline{\noalign{\global\savewidth\arrayrulewidth
  \global\arrayrulewidth 1pt}\hline\noalign{\global\arrayrulewidth\savewidth}}

\usepackage{orcidlink}

\def\wacvPaperID{*****} 
\def\confName{WACV}
\def\confYear{2025}

\begin{document}

\title{PVP: Polar Representation Boost for 3D Semantic Occupancy Prediction}

\author{Yujing Xue*\\
NUS\\
{\tt\small xueyj14@outlook.com}
\and
Jiaxiang Liu*\\    
Zhejiang University\\
{\tt\small jiaxiang.21@intl.zju.edu.cn}
\and
Jiawei Du\dag \\
A*STAR, NUS\\
{\tt\small dujiawei@u.nus.edu}
\and
Joey Tianyi Zhou\\
A*STAR\\
{\tt\small joey.tianyi.zhou@gmail.com}
}

\maketitle

\begin{abstract}
\vspace{-1em}
Recently, representations based on polar coordinates have exhibited promising characteristics for 3D perceptual tasks.
In addition to Cartesian-based methods, representing surrounding spaces through polar grids offers a compelling alternative in these tasks. This approach is advantageous for its ability to represent larger areas while preserving greater detail of nearby spaces. 
However, polar-based methods are inherently challenged by the issue of feature distortion due to the non-uniform division inherent to polar representation.
To harness the advantages of polar representation while addressing its challenges, we propose \textbf{P}olar \textbf{V}oxel
Occupancy \textbf{P}redictor (PVP), a novel 3D multi-modal occupancy predictor operating in polar coordinates.
PVP mitigates the issues of feature distortion and misalignment across different modalities with the following two design elements:
1) Global Represent Propagation (GRP) module, which incorporates global spatial information into the intermediate 3D volume, taking into account the prior spatial structure. It then employs Global Decomposed Attention to accurately propagate features to their correct locations.
2) Plane Decomposed Convolution (PD-Conv), which simplifies 3D distortions in polar coordinates by replacing 3D convolution with a series of 2D convolutions.
With these straightforward yet impactful modifications, our PVP surpasses state-of-the-art works by significant margins—improving by 1.9\% mIoU and 2.9\% IoU over LiDAR-only methods, and by 7.9\% mIoU and 6.8\% IoU over multimodal methods on the OpenOccupancy dataset.
\let\thefootnote\relax\footnotetext{
*Contributed equally.
\dag Corresponding author.}
\end{abstract}

\section{Introduction}\label{1}
In autonomous driving systems, accurately perceiving 3D structural information about surrounding scenes, including both objects and regions, is critically important, especially when High Definition maps are unavailable \cite{pointocc,pan20213d,zhao2023lif}.
Previous tasks like 3D object detection and 3D semantic segmentation have mainly focused on foreground objects or scanned points \cite{nie2023partner,liu2023deep,pan20213d,yin2021center,graham20183d,tchapmi2017segcloud,li2023mseg3d,meng2023hydro,qin2023supfusion,hao2022ai,zheng2024refined}.
This means focusing too narrowly limits a system's ability to fully understand diverse shapes in its environment, reducing its adaptability to new objects.

To overcome these limitations, the task of 3D semantic occupancy  prediction has been proposed, aiming to assign a semantic label to every voxel within the perceptual range.
This task is gaining increased attention and is becoming a crucial element in 3D scene understanding for autonomous driving.
Previous methods~\cite{huang2023tri, zhang2023occformer, wang2023openoccupancy, wei2023surroundocc,zhang2023scalable,tang2021sa} have employed various techniques to capture the appropriate representative features for each voxel. For instance, 
TPVFormer~\cite{huang2023tri} models the surrounding space using a tri-plane approach and employs sparse sampling to guide the voxel prediction head. This demonstrates the potential of 2D representation and sampling-based supervision.
SurroundOcc~\cite{wei2023surroundocc} and BEVFormer~\cite{li2022bevformer} utilize a set of spatial queries along with multi-scale feature aggregation to derive the spatial volume feature. They then apply a 3D convolutional network to extract the features.
OccFormer~\cite{zhang2023occformer} and CONet~\cite{wang2023openoccupancy} utilize the lift-splat-shoot pipeline to generate spatial volume features. Beside, OccFormer~\cite{zhang2023occformer} enhances this approach by incorporating additional Bird's Eye View (BEV) features and a window-attention-based Feature Pyramid Network (FPN) \cite{lin2017feature} with cross-attention, leveraging the extra BEV features for improved performance.
CONet~\cite{wang2023openoccupancy} is the pioneering work that introduced a modern multi-modality pipeline and employed direct sampling to obtain the prediction from a volume down-sampled by a factor of 4.
Nevertheless, the aforementioned most methods rely on Cartesian-based volumetric representation, which leads to a uniform voxel distribution across distances. 
This uniformity leads to distant regions having the same voxel density as nearby areas, even though the information available, such as points or visual cues in images for these distant regions, is diminished due to the inherent limitations of sensors. Consequently, this situation results in computational waste and a scarcity of information for distant region voxel \cite{nie2023partner}.
Therefore, Polar representation has received widespread attention because it demonstrates the ability to alleviate this problem~\cite{nie2023partner,jiang2023polarformer,kim2023craft,jhaldiyal2023semantic,feng2024polarpoint}.

In polar coordinates, the volume representation of point clouds, which varies with distance, naturally aligns with the uneven distribution of geometric information from point clouds and semantic information from cameras.
Moreover, the success of polar representation in LiDAR semantic segmentation tasks~\cite{zhang2020polarnet, zhu2021cylindrical} and 3D object detection~\cite{nie2023partner} motivates us to explore polar representation in 3D occupancy prediction.
However, despite its promising characteristics, a polar-based 3D occupancy predictor inevitably encounters severe feature distortions \cite{nie2023partner}.
As discussed in previous research \cite{nie2023partner}, 
scenes represented by polar grids appear distorted according to their range, as shown in \autoref{distortion}. This distortion introduces two main challenges in polar representation:
1) In polar grid representations, the distinct and varying occupancy patterns of local structures make the translation-invariant convolution poorly adapted for non-rectangular geometries.
This incompatibility leads to the extraction of unsuitable local features, thereby compromising the efficacy of polar-based predictive models.
2) Entities such as roads, sidewalks, or ground surfaces are subject to pronounced distortions in their global structural integrity. This distortion poses a significant challenge to the process of feature propagation, as it transitions from sparse feature grids to more dense representations. Local feature propagators, including convolutional operations or window-based attention mechanisms, may fall short in capturing and understanding the comprehensive global structure of these object classes. Consequently, this limitation hampers the accurate dissemination of features across the corresponding voxels.
Although PolarStream~\cite{chen2021polarstream} attempted to tackle this issue by using range-stratified convolutions, its complex design falls short in realigning features for effective object recognition. 
{PARTNER~\cite{nie2023partner} introduces two mechanisms aimed at realigning features of foreground objects for feature distortion; however, this design is specifically tailored for foreground objects and does not address the broader issue of feature propagation.}

To tackle the aforementioned challenges, we propose a novel \textbf{P}olar \textbf{V}oxel
Occupancy \textbf{P}redictor, abbreviated as PVP, designed to learn effective representations in polar coordinates.
We design two key components: the global representation propagation module (\textbf{GRP} Module) and plane decomposed convolution (\textbf{PD-Conv}):
1) {The \textbf{GRP} module utilizes a novel attention mechanism to capture the road structure from the scene volume and accurately propagate features to their correct locations. To prevent the introduction of burdensome global attention, we initially compress the input volume using window attention to extract representative features. Following the condensation of features within windows, dimension-wise self-attention is applied for effective global feature propagation.}
2) To mitigate local distortion inherent to polar volume representations, we design \textbf{PD-Conv}. This approach simplifies the handling of complex 3D distortions by replacing traditional 3D convolutions with three distinct 2D convolutions. These convolutions correspond to a scale transformation on the range plane, a projection transformation on the BEV plane, and an identity transformation on the slicing plane, effectively addressing the distortion challenges specific to polar volumes.
Armed with the proposed designs, our PVP outperforms existing methods, achieving superior performance, as shown in \autoref{ECCV-Figure1}. Our contributions can be summarized as follows:

\begin{figure}[t]\scriptsize
	\begin{center}
		\includegraphics[width=\linewidth]{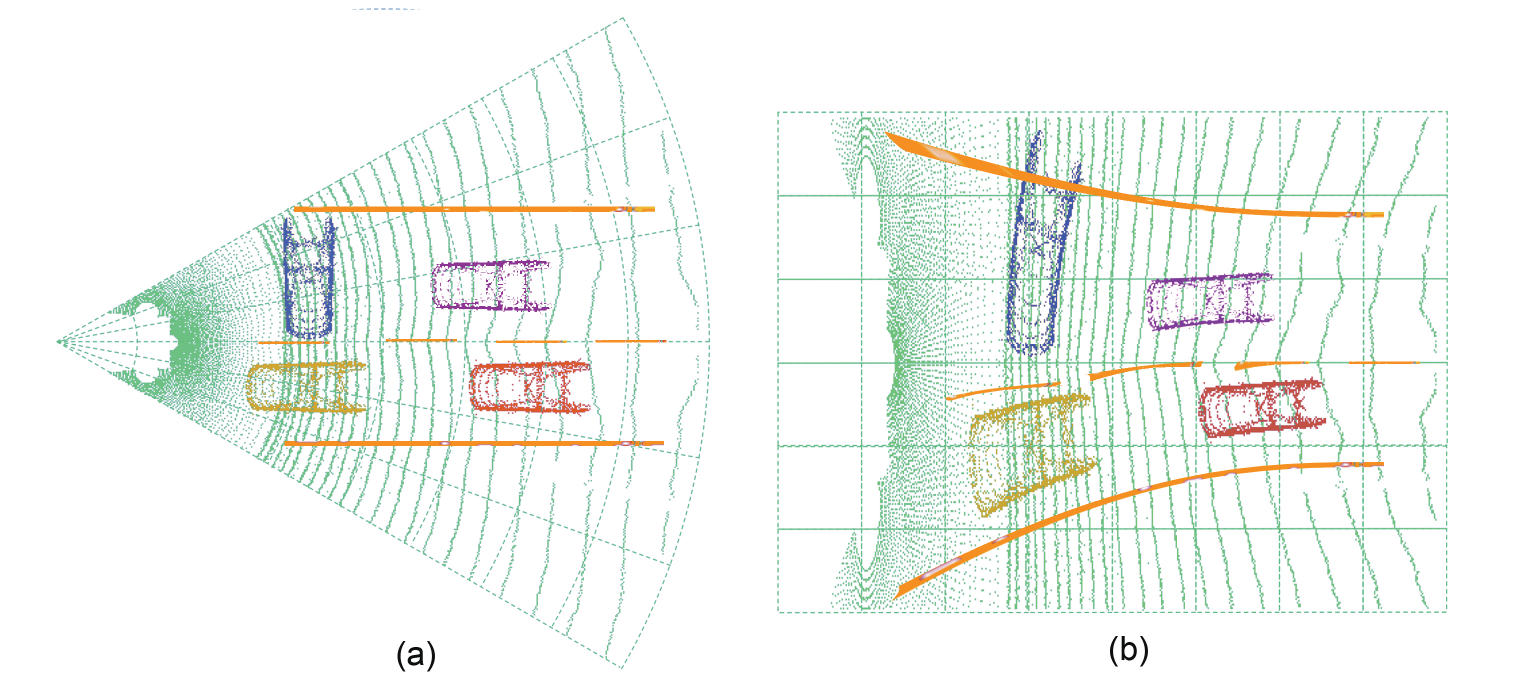}
	\end{center}
	\caption{An illustration of polar feature distortion: The non-uniform division of polar representation causes identical objects (e.g., cars) and scenes (e.g., lane lines) at different ranges and headings to exhibit varied distorted appearances. This leads to global misalignment between objects and scenes and increases the challenge of regression for polar-based Occupancy Prediction.
    }
	\label{distortion}
\end{figure}
\begin{figure}[t]\scriptsize
	\begin{center}
		\includegraphics[width=\linewidth]{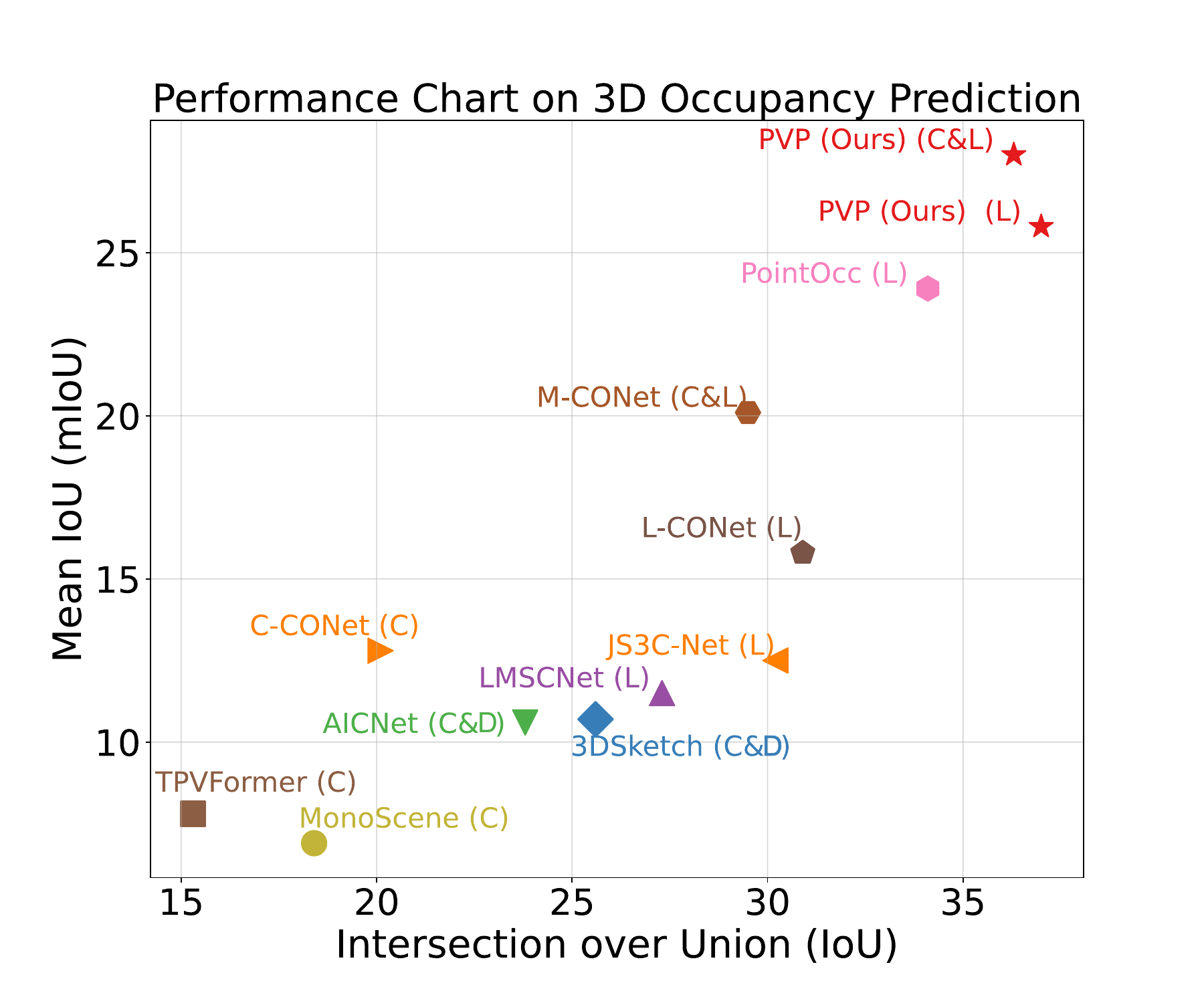}
	\end{center}
  \vspace{-2em}
	\caption{Performance Chart on 3D Occupancy Prediction: Our method \textit{PVP} achieved the best results. The Intersection over Union (IoU), acting as the geometric metric to distinguish whether a voxel is occupied or empty (with all occupied voxels considered as one category), and the mean IoU (mIoU) across all classes, serving as the semantic metric, were utilized. C\&L signifies that the input includes both camera images and LiDAR point clouds. L indicates that the input is exclusively LiDAR point clouds. C represents that the input contains only camera images. C\&D means the input comprises both camera images and depth images.}
	\label{ECCV-Figure1}
\end{figure}


\begin{itemize}
\item We investigate the feature distortion issue in polar volume representation and, to address this challenge, we specifically propose a novel 3D occupancy predictor, dubbed PVP,  featuring GRP and PD-Conv key designs.

\item The GRP module designs a attention mechanism to capture road structure and accurately propagate features by initially compressing the input volume with window attention, then applying global decomposed attention, effectively addressing feature localization without burdensome global attention.

\item To mitigate local distortion in polar volume representations, PD-Conv replaces traditional 3D convolutions with distinct 2D convolutions, addressing complex 3D distortions through scale, projection, and identity transformations on different planes, simplifying the handling of these distortions.


\item We extensively validate the effectiveness of our designs on the OpenOccupancy dataset. PVP achieves remarkable performance gains over state-of-the-art (SOTA) methods on both geometric and semantic metrics, showcasing superior benefits for inputs from both LiDAR and multi-modal sources.

\end{itemize}



\section{Related work}

\subsection{Polar-based 3D Perception Tasks}
Polar or polar-like coordinate representations have been extensively explored in several previous tasks.
PolarNet~\cite{zhang2020polarnet} and Cylinder3D~\cite{zhu2021cylindrical} employ polar-shaped partitions to evenly distribute points across grid cells in LiDAR segmentation tasks.
PolarStream~\cite{chen2021polarstream} further deploys polar representation to 3D detection by proposing a Polar-to-Cartesian sampling module and a range-stratified convolution.
OT \cite{yang2023one} employs Polar representation for adaptive, resolution-flexible BEV feature mapping, enabling efficient, one-time training for diverse deployments in 3D perception.
{PARTNER~\cite{nie2023partner} incorporates both long-range and local attention mechanisms into 3D detection to realign the polar features.}
However, previous methods have not addressed the challenges posed by dense voxel prediction. The use of 3D polar representation exacerbates the distortion issue compared to BEV or sparse 3D representations, particularly in terms of feature propagation.



\subsection{3D Occupancy Prediction}
Previous tasks such as LiDAR segmentation \cite{zhang2020polarnet, zhu2021cylindrical,zhao2023lif,li2023mseg3d} and object detection~\cite{nie2023partner,meng2023hydro,qin2023supfusion} have focused on labeling only a limited number of sparse LiDAR points or foreground objects. These approaches tend to overlook a comprehensive representation of the surrounding environment and demonstrate limited generalization to objects with unfamiliar geometries.
3D occupancy prediction~\cite{wang2023openoccupancy, behley2019semantickitti} seeks to assign a semantic label to every voxel within the perceptual range, thus making it a cornerstone for 3D scene comprehension in autonomous driving.
In 3D occupancy prediction task, the requirement for dense prediction space and cubic voxel labeling has led most existing works ~\cite{li2022bevformer,wei2023surroundocc, roldão2020lmscnet, yan2020sparse, li2020anisotropic} to adopt a Cartesian-based voxel model .
Cartesian-based voxel models, despite their efficacy, uniformly distribute voxels across distances. This results in equal voxel density for both distant and nearby regions, reducing the detail for distant areas due to sensor limitations.
Therefore, methods based on polar coordinates hold significant potential in the 3D occupancy prediction task.
The potential stems from its adept alignment with the uneven distribution of geometric and semantic information characteristic of point clouds and camera feeds.


\begin{figure}[t]\scriptsize
	\begin{center}
		\includegraphics[width=1.0\linewidth]{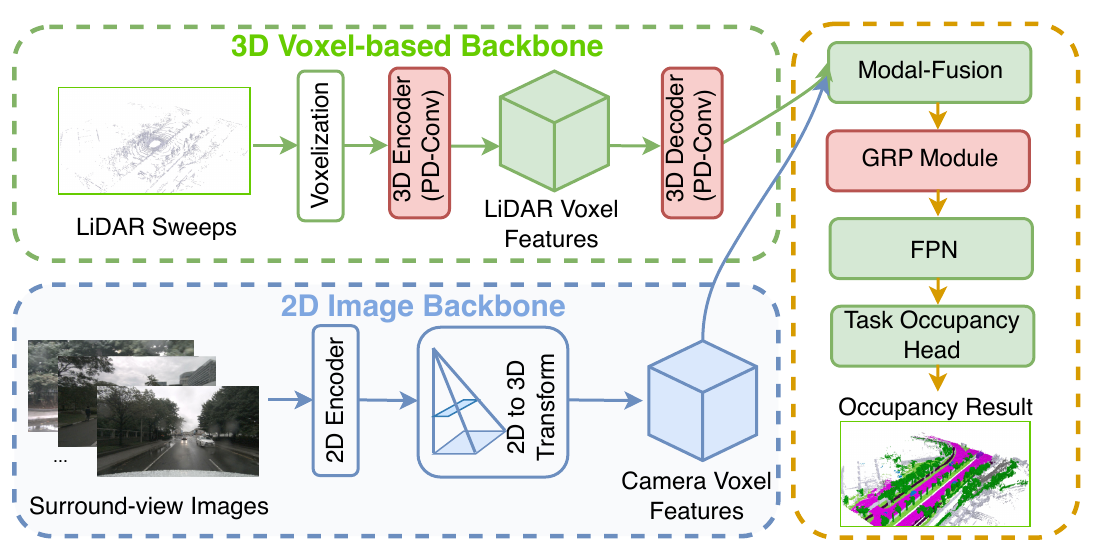}
	\end{center}
  \vspace{-2em}
	\caption{The pipeline of \textit{PVP}. Our proposed PVP consists of three components: 1) The grid-based feature extraction and fusion module, which includes a 3D Voxel-based Backbone with PD-Conv for feature extraction and a 3D Image backbone for 2D to 3D feature conversion. 2) The GRP Module utilizes attention mechanisms to capture road structures from the scene volume and accurately propagate features to their correct locations. 3) The FPN processes dense features for feature aggregation, followed by the task occupancy head. This head comprises an occupancy head and a point cloud refinement module. The primary goal of the task occupancy head is to transform the polar-based 3D tensor into Cartesian voxel output for enhanced results.
    }
	\label{Pipeline}
 \vspace{-2em}
\end{figure}
\section{Method}
This section introduces PVP, a dense voxel prediction model designed for polar representation. The overall architecture is introduced in Section~\ref{Overall Architecture}, as depicted in \autoref{Pipeline}. 
Then, we elaborate on the two key techniques of PVP: the GRP Module in Section~\ref{grpmodule} and PD-Conv in Section~\ref{pdconv}.
Finally we discuss the design of task head and losses in Section~\ref{headnloss}
\subsection{Overview} \label{Overall Architecture}
For each 3D scene, \textit{PVP} processes input comprising a LiDAR point cloud with $N_p$ points and several images capturing the surrounding environment.
Each point is represented by a feature vector $(r_p, \theta_p, x_p, y_p, z_p, i_p)$, where $(r_p, \theta_p)$ is the polar coordinates, $(x_p, y_p, z_p)$ is the Cartesian coordinates, and $i_p$ is the reflection intensity \cite{wang2023openoccupancy}.
Our proposed \textit{PVP} consists of three components:

1) \textit{The 2D/3D features extraction and fusion module}. Inspired by Openoccupancy \cite{wang2023openoccupancy}, PVP includes two backbones. 
For 3D voxel-based backbone, 3D voxel-based backbone, consisting of PD-Conv, leverages grid feature embeddings from points in voxels or pillars and compresses sparse point features into a dense 3D feature tensor in polar coordinates. Specifically,
the raw LiDAR sweep points are first embedded into voxelized features through parameterized voxelization, encoded and dimensionality-reduced by 3D sparse convolutions, then decoded by to generate dense voxel features $F^\mathcal{L}$.
For a 2D image backbone, recent works~\cite{li2022bevformer, philion2020lift} trying to lift features from images to 3D space in a spatial projection manner. 
Our 2D image backbone also extracts semantic features that are projected into 3D spaces for enhanced semantic performance. Specifically, for surround-view images, multi-view features are extracted using a 2D encoder, such as ResNet \cite{he2016deep}. Then, to fully leverage the semantic features from images, we apply a 2D to 3D transform \cite{liu2023bevfusion} to project 2D features into 3D coordinates, with preserved height information for 3D occupancy prediction. This process yields camera voxel features \(F^\mathcal{C}\) \cite{wang2023openoccupancy}.

The 3D features \( F^\mathcal{L} \) and 2D camera voxel features \( F^\mathcal{C} \) are natural representations for 3D semantic occupancy prediction. The Modal-Fusion Module takes the 3D tensor produced by both 3D backbone and 2D backbone as input and produces a fused 3D tenser $F^\mathcal{F}$ \cite{wang2023openoccupancy}:

\begin{equation}
F^\mathcal{F} = W \odot F^\mathcal{L} + (1 - W) \odot F^\mathcal{C},
\end{equation}

where \(\odot\) represents element-wise product. The calculation of \(W\) follows the methodology described in reference \cite{wang2023openoccupancy}. 

2) \textit{GRP Module.} GRP takes the $F^\mathcal{F}$  as input and produces a dense 3D tenser by introducing attention for global feature propagation. The technical details are provided in Section \ref{grpmodule}.

3) \textit{FPN and polar prediction head.} 
FPN \cite{lin2017feature} processes the dense features for feature aggregation, and then the polar prediction head, which includes occupancy head and a point cloud refinement module \cite{wang2023openoccupancy}. The goal of the polar prediction head is to transform the polar-based 3D tensor into Cartesian voxel output to achieve better results.


\subsection{Global Representation Propagation} \label{grpmodule}
\begin{figure*}[t]\scriptsize
	\begin{center}
		\includegraphics[width=1.0\linewidth]{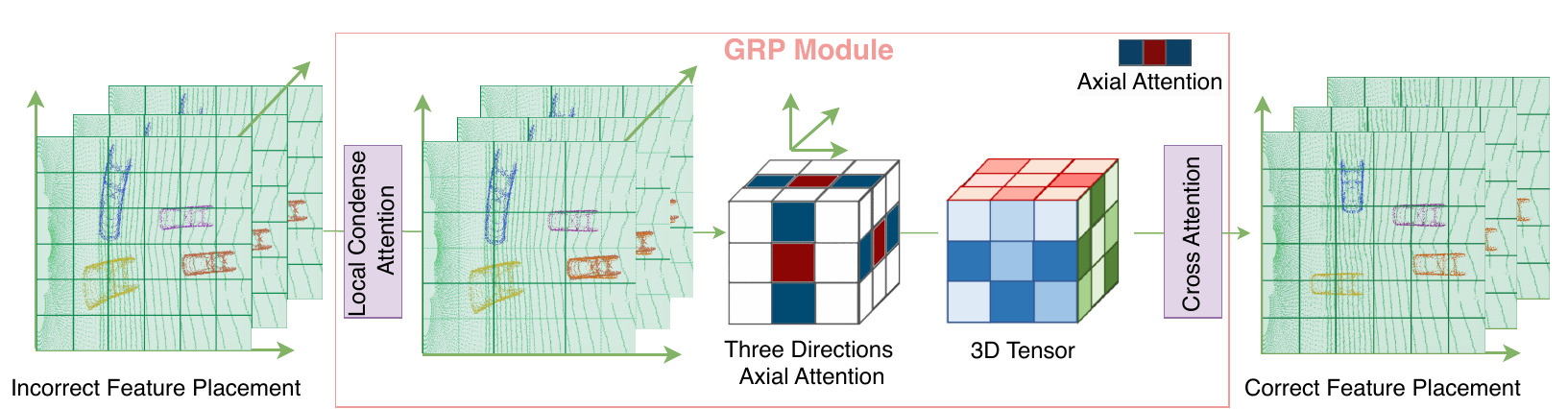}
	\end{center}
  \vspace{-2em}
	\caption{GRP module encompasses two types of attention sub-modules: 1) Local condense attention for condensing multi-modal local features, and 2) Global decomposed attention, which leverages axial attention across three directions for enhanced feature extraction, followed by cross-attention for the recalibration of long-range feature propagation. After GRP module, the distorted features are corrected.
    }
	\label{ECCV-Figure-GRP}
\end{figure*}

In this section, we elaborate on the GRP module, which applies attention to essential representative features for global calibration, as shown in \autoref{ECCV-Figure-GRP}.
The design of GRP module is inspired by two key observations: (1) the spatial resolution of polar grids varies with distance, and (2) the input 3D tensors are typically very sparse. Due to these characteristics, structural elements such as roads, sidewalks, trees, or large buildings often experience significant gaps and distortions in their occupied regions. These issues lead to a noticeable decline in performance for such classes and negatively affect the performance of CNN-based predictors, particularly in terms of geometric accuracy, given the significant presence of these classes.
{To address these challenges, the GRP employs two types of attention sub-modules: \underline{local condense attention} for multi-modal local feature condensation and \underline{global decomposed attention} for the re-calibration of long-range feature propagation. The specifics of these two sub-modules will be detailed subsequently.}
\subsubsection{Local Condense Attention.} 
The proposed local condense attention takes the 3D feature $F^\mathcal{F} \in \mathbb{R}^{R \times A \times Z \times C}$ as input, where $R$, $A$, and $Z$ denote the resolution of the radial, azimuth and height space, and $C$ denotes the number of feature channels.
To select representative features and reduce the computational cost of subsequent steps, we introduce a local window-based max-selection operation. We first divide $F^\mathcal{F}$ into non-overlapping windows with size $W_g \times W_g \times W_g$. Let $F^\mathcal{F}[i] \in \mathbb{R}^{S \times S \times S \times C}$ be the $i$-th local window, the represent feature of the window $f_i$ can be selected as following: $ f_{i} = \mathop{maxsel}(F^\mathcal{F}[i])$,
where $\mathop{maxsel}$ is a max selection operation that filters out local-maxima in the local window $f_i \in \mathbb{R}^{C}$. 
The suppression operation retains the most representative features within each window. Subsequently, we compress each window to its representative features, resulting in a more compact global scene representation.
Subsequently, we compress each window to its representative features.
Let $f_{i}$ and $F^\mathcal{F}[i]$ be the query and attending features of the $i$-th window, and $p_{i}, p^{'}_{i}$ be the corresponding coordinates of the real pixel centers in both Polar and Cartesian systems.
For the $i$-th window, we conduct dot-product attention as follows:
 \begin{equation}
    \begin{split}
    \centering
             f^{rep}_{i} &= softmax(\frac{Q_{i}K_{i}}{\sqrt{d}} \cdot V_{i}+E(p)), \quad \mbox{where} \\
         Q_{i} &= f_{i}W_{q}, \quad K_{i} = F^\mathcal{F}[i]W_{k}, \quad V_{i} = F^\mathcal{F}[i]W_{v},\\
    \end{split}
    \label{3.2.3}
\end{equation}
$W_{q}, W_{k}, W_{v}$ are the linear projection of query, key, and value, and $E(p)$ is a function calculating the relative positional encoding as:
\begin{equation} \label{3.2.4}
    E(p) = ReLU((p_i - p^{'}_{i}) \cdot W_{pos}).
\end{equation}
The local condense attention is executed for each window to get a condensed feature map $F^{rep} = (f^{rep}_{1}, \cdots, f^{rep}_{n}) \in \mathbb{R}^{\frac{R}{S} \times \frac{A}{S} \times \frac{Z}{S} \times C}$ for the following global decomposed attention.

\subsubsection{Global Decomposed Attention.}
Leveraging the condensed representative features obtained from local cross attention, the global decomposed attention mechanism employs long-range self-attention for the propagation and recalibration of global features.
The design is informed by the observation that stuff classes typically span the entire scene, and different parts of a structural element may experience varying levels of distortion. Therefore, it is crucial to implement global spatial attention, enabling distant distorted features to interact without significantly increasing computational demands.
Building on this understanding, we introduce decomposed axis window attention to implement lightweight and efficient global attention without compromising spatial structure information. We use the attention mechanism along the radial axis as an example to illustrate the details of our design.
The condensed feature map $F^{rep} \in \mathbb{R}^{\frac{R}{S} \times \frac{A}{S} \times \frac{Z}{S} \times C}$ is first divided into non-overlapping windows as strips along the radial axis, where the window size is set to $\frac{R}{S} \times 1 \times 1$. For feature $F^{rep}[i]$ in the $i$-th window, self-attention is applied following Eq.~\ref{3.2.3}-\ref{3.2.4}.
Our model employs three stacked modules for information interaction, where the following second and third decomposed attention uses window partitions along the azimuth and height axes.
After decomposed attention, the module applies reverse local cross attention (Eq. \ref{3.2.3}-\ref{3.2.4}) to propagate the representative features to each window of the feature map.
Let $F^{r}$ denote the output feature of the last decomposed attention \cite{nie2023partner}.
With $f_{i}= F[:,i] \in \mathbb{R}^{R \times C}$ and $f^{'}_{i} = F^{r}[:,i] \in \mathbb{R}^{N \times C}$ as the query and attending features, respectively.

\subsection{Plane Decomposed Convolution} \label{pdconv}
{In this section, we reexamine the issue of 3D spatial distortion in polar representation. Traditional 3D convolutions struggle with the unbalanced distribution, leading to compromised feature representation. However, the distortion of 3D shapes in polar coordinates varies across different views due to the asymmetric partitioning of axes (range, azimuth, and height).}
Most prominently, the distortion can be categorized into 
projection distortion in the top view $F_{(r, \theta)}$, and scale distortion in the range view $F_{(r, z)}$.
Meanwhile, it remains unchanged in the slice view $F_{(\theta, z)}$.

\begin{figure}[t]\scriptsize
	\begin{center}
		\includegraphics[width=\linewidth]{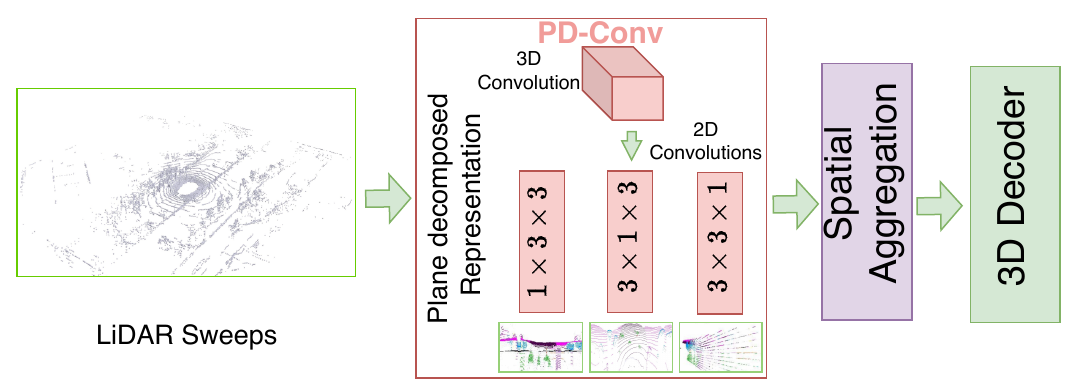}
	\end{center}
	\caption{A diagram illustrating PD-Conv. PD-Conv simplifies the handling of complex 3D distortions by substituting traditional 3D convolutions with three separate 2D convolutions. These convolutions facilitate a scale transformation on the range plane, a projection transformation on BEV plane, and an identity transformation on the slicing plane, effectively addressing the local distortion typical of polar volume representations.
    }
	\label{ECCV-Figure-PD-Conv}
\end{figure}

Recent work~\cite{pointocc} has also observed this phenomenon and attempted to address it by proposing a tri-planar projection method.
Tri-planar projection allows the 3D space to be projected into three different views, enabling the processing of point cloud features using 2D convolutions. After processing in these 2D views, the features are aggregated to reconstruct the 3D representation. While this approach offers some improvements, the loss of spatial structure renders the method less effective.
To maintain spatial structure, we have developed the PD-Conv, a technique that preserves the point cloud representation in 3D tensors during the entire processing phase, as shown in \autoref{ECCV-Figure-PD-Conv}.
\subsubsection{Decomposed Block}
We start by presenting our single decomposed block, followed by detailing the concatenated structure of the PD-Conv.
In decomposed blocks, we substitute traditional 3D convolutional kernels with asymmetric convolutional kernels~\cite{zhu2021cylindrical}.
Specifically, for a 3D convolution with $3\times3\times3$ kernels, in which the axis represents range, azimuth and height, we decompose a single axis to obtain three different decomposed kernels: $1 \times 3 \times 3$, $3 \times 1 \times 3$ and $3 \times 3 \times 1$.

\subsubsection{Plane Decomposed Modeling}
After substituting the 3D convolutional kernels with asymmetric decomposed ones, we stack various decomposed blocks to ensure a robust spatial representation. The stacking approach can be serial, parallel, or a combination of both. Experimental verification has shown that the stacking mode does not significantly affect performance.
In this paper, we opt to stack three distinct decomposed blocks in series ($1 \times 3 \times 3$, $3 \times 1 \times 3$ and $3 \times 3 \times 1$), generating different stacks by varying their order, and then combine them in parallel. 
These convolutions facilitate specific transformations: a scale transformation on the range plane, a projection transformation on the BEV plane, and an identity transformation on the slicing plane. 
After implementing the plane decomposed representation and spatial aggregation, feature distortion in 3D space is efficiently separated across different planes. This process ensures the preservation of spatial structure, effectively tackling the distortion issues unique to polar volumes.

\subsection{Task Occupancy Head} \label{headnloss}
In this section, we detail the design of the polar prediction head and the loss function. Upon obtaining the final polar feature grids \(P_{final}\), we employ a straightforward sampling strategy to generate Cartesian outputs. The process begins with a coordinate transformation to determine the center coordinate of the output Cartesian voxel in Polar representation \cite{fan2021scf}:
\begin{equation} 
    (r, \theta, z) = ((x^2 + y^2)^\frac{1}{2}, atan(y, x), z)
\end{equation}
Then, for a voxel with center coordinates $(r_v, \theta_v, z_v)$, we perform trilinear sampling $\mathcal{S}$ to obtain its feature 
 $f_v$:
\begin{equation} 
    f_v = \mathcal{S}(P_{final}, (r_v, \theta_v, z_v)),
\end{equation}
After obtaining the Cartesian feature grid, we apply the same heads, a coarse-to-fine strategy, and training loss as described in \cite{wang2023openoccupancy} to produce the final output.

\definecolor{LightGrey}{rgb}{.9,.9,.9}
\definecolor{White}{rgb}{1.,0.,1.}
\definecolor{first}{rgb}{.8,.0,.0}
\definecolor{second}{rgb}{.0,.6,.0}
\definecolor{third}{rgb}{.0,.0,.8}

\definecolor{nbarrier}{RGB}{255, 120, 50}
\definecolor{nbicycle}{RGB}{255, 192, 203}
\definecolor{nbus}{RGB}{255, 255, 0}
\definecolor{ncar}{RGB}{0, 150, 245}
\definecolor{nconstruct}{RGB}{0, 255, 255}
\definecolor{nmotor}{RGB}{200, 180, 0}
\definecolor{npedestrian}{RGB}{255, 0, 0}
\definecolor{ntraffic}{RGB}{255, 240, 150}
\definecolor{ntrailer}{RGB}{135, 60, 0}
\definecolor{ntruck}{RGB}{160, 32, 240}
\definecolor{ndriveable}{RGB}{255, 0, 255}
\definecolor{nother}{RGB}{139, 137, 137}
\definecolor{nsidewalk}{RGB}{75, 0, 75}
\definecolor{nterrain}{RGB}{150, 240, 80}
\definecolor{nmanmade}{RGB}{213, 213, 213}
\definecolor{nvegetation}{RGB}{0, 175, 0}

\definecolor{car}{rgb}{0.39215686, 0.58823529, 0.96078431}
\definecolor{bicycle}{rgb}{0.39215686, 0.90196078, 0.96078431}
\definecolor{motorcycle}{rgb}{0.11764706, 0.23529412, 0.58823529}
\definecolor{truck}{rgb}{0.31372549, 0.11764706, 0.70588235}
\definecolor{other-vehicle}{rgb}{0.39215686, 0.31372549, 0.98039216}
\definecolor{person}{rgb}{1.        , 0.11764706, 0.11764706}
\definecolor{bicyclist}{rgb}{1.        , 0.15686275, 0.78431373}
\definecolor{motorcyclist}{rgb}{0.58823529, 0.11764706, 0.35294118}
\definecolor{road}{rgb}{1.        , 0.        , 1.        }
\definecolor{parking}{rgb}{1.        , 0.58823529, 1.        }
\definecolor{sidewalk}{rgb}{0.29411765, 0.        , 0.29411765}
\definecolor{other-ground}{rgb}{0.68627451, 0.        , 0.29411765}
\definecolor{building}{rgb}{1.        , 0.78431373, 0.        }
\definecolor{fence}{rgb}{1.        , 0.47058824, 0.19607843}
\definecolor{vegetation}{rgb}{0.        , 0.68627451, 0.        }
\definecolor{trunk}{rgb}{0.52941176, 0.23529412, 0.        }
\definecolor{terrain}{rgb}{0.58823529, 0.94117647, 0.31372549}
\definecolor{pole}{rgb}{1.        , 0.94117647, 0.58823529}
\definecolor{traffic-sign}{rgb}{1.        , 0.        , 0.    }   

\makeatletter
\newcommand{\car@semkitfreq}{3.92}
\newcommand{\bicycle@semkitfreq}{0.03}
\newcommand{\motorcycle@semkitfreq}{0.03}
\newcommand{\truck@semkitfreq}{0.16}
\newcommand{\othervehicle@semkitfreq}{0.20}
\newcommand{\person@semkitfreq}{0.07}
\newcommand{\bicyclist@semkitfreq}{0.07}
\newcommand{\motorcyclist@semkitfreq}{0.05}
\newcommand{\road@semkitfreq}{15.30}  %
\newcommand{\parking@semkitfreq}{1.12}
\newcommand{\sidewalk@semkitfreq}{11.13}  %
\newcommand{\otherground@semkitfreq}{0.56}
\newcommand{\building@semkitfreq}{14.1}  %
\newcommand{\fence@semkitfreq}{3.90}
\newcommand{\vegetation@semkitfreq}{39.3}  %
\newcommand{\trunk@semkitfreq}{0.51}
\newcommand{\terrain@semkitfreq}{9.17} %
\newcommand{\pole@semkitfreq}{0.29}
\newcommand{\trafficsign@semkitfreq}{0.08}
\newcommand{\semkitfreq}[1]{{\csname #1@semkitfreq\endcsname}}

\newcommand{\barrier@nuscenesfreq}{11.79}
\newcommand{\bicycle@nuscenesfreq}{0.18}
\newcommand{\bus@nuscenesfreq}{5.83}
\newcommand{\car@nuscenesfreq}{48.27}
\newcommand{\construction@nuscenesfreq}{1.92}
\newcommand{\motorcycle@nuscenesfreq}{0.54}
\newcommand{\pedestrian@nuscenesfreq}{2.93}
\newcommand{\trafficcone@nuscenesfreq}{0.93}
\newcommand{\trailer@nuscenesfreq}{6.22}
\newcommand{\truck@nuscenesfreq}{20.07}
\newcommand{\driveable@nuscenesfreq}{28.64}
\newcommand{\other@nuscenesfreq}{0.77}
\newcommand{\sidewalk@nuscenesfreq}{6.34}
\newcommand{\terrain@nuscenesfreq}{6.35}
\newcommand{\manmade@nuscenesfreq}{16.10}
\newcommand{\vegetation@nuscenesfreq}{11.08}
\newcommand{\nuscenesfreq}[1]{{\csname #1@nuscenesfreq\endcsname}}

\begin{table*}[t] 
	\setlength{\tabcolsep}{0.0085\linewidth}
 \setlength{\tabcolsep}{1pt} 
    \renewcommand\arraystretch{1.3}
	\caption{\textbf{3D Semantic occupancy prediction results on nuScenes validation set~\cite{caesar2020nuscenes}.} The C, L, and D denotes camera, LiDAR, and depth, respectively. Our PVP achieves better performance than all previous methods based on all input modalities.}
	
	\centering
	\begin{tabular}{ c| c | c c | c c c c c c c c c c c c c c c c}
		\toprule
		Method & Input & IoU & mIoU
		& \rotatebox{90}{\textcolor{nbarrier}{$\blacksquare$} barrier}
		& \rotatebox{90}{\textcolor{nbicycle}{$\blacksquare$} bicycle}
		& \rotatebox{90}{\textcolor{nbus}{$\blacksquare$} bus}
		& \rotatebox{90}{\textcolor{ncar}{$\blacksquare$} car}
		& \rotatebox{90}{\textcolor{nconstruct}{$\blacksquare$} const. veh.}
		& \rotatebox{90}{\textcolor{nmotor}{$\blacksquare$} motorcycle}
		& \rotatebox{90}{\textcolor{npedestrian}{$\blacksquare$} pedestrian}
		& \rotatebox{90}{\textcolor{ntraffic}{$\blacksquare$} traffic cone}
		& \rotatebox{90}{\textcolor{ntrailer}{$\blacksquare$} trailer}
		& \rotatebox{90}{\textcolor{ntruck}{$\blacksquare$} truck}
		& \rotatebox{90}{\textcolor{ndriveable}{$\blacksquare$} drive. suf.}
		& \rotatebox{90}{\textcolor{nother}{$\blacksquare$} other flat}
		& \rotatebox{90}{\textcolor{nsidewalk}{$\blacksquare$} sidewalk}
		& \rotatebox{90}{\textcolor{nterrain}{$\blacksquare$} terrain}
		& \rotatebox{90}{\textcolor{nmanmade}{$\blacksquare$} manmade}
		& \rotatebox{90}{\textcolor{nvegetation}{$\blacksquare$} vegetation}
		\\
            \midrule
            \midrule

        MonoScene~\cite{cao2022monoscene} & C & 18.4 & 6.9 & 7.1 & 3.9 & 9.3 & 7.2 & 5.6 & 3.0 & 5.9 & 4.4 & 4.9 & 4.2 & 14.9 & 6.3 & 7.9 & 7.4 & 10.0 & 7.6 \\

        TPVFormer~\cite{huang2023tri} & C & 15.3 & 7.8 & 9.3 & 4.1 & 11.3 & 10.1 & 5.2 & 4.3 & 5.9 & 5.3 & 6.8 & 6.5 & 13.6 & 9.0 & 8.3 & 8.0 & 9.2 & 8.2 \\

        3DSketch~\cite{chen20203d} & \makecell{C\&D} & 25.6 & 10.7 & 12.0 & 5.1 & 10.7 & 12.4 & 6.5 & 4.0 & 5.0 & 6.3 & 8.0 & 7.2 & 21.8 & 14.8 & 13.0 & 11.8 & 12.0 & 21.2 \\

        AICNet~\cite{li2020anisotropic} & \makecell{C\&D} & 23.8 & 10.6 & 11.5 & 4.0 & 11.8 & 12.3 & 5.1 & 3.8 & 6.2 & 6.0 & 8.2 & 7.5 & 24.1 & 13.0 & 12.8 & 11.5 & 11.6 & 20.2 \\

        LMSCNet~\cite{roldão2020lmscnet} & L & 27.3 & 11.5 & 12.4 & 4.2 & 12.8 & 12.1 & 6.2 & 4.7 & 6.2 & 6.3 & 8.8 & 7.2 & 24.2 & 12.3 & 16.6 & 14.1 & 13.9 & 22.2 \\

        JS3C-Net~\cite{yan2020sparse} & L & 30.2 & 12.5 & 14.2 & 3.4 & 13.6 & 12.0 & 7.2 & 4.3 & 7.3 & 6.8 & 9.2 & 9.1 & 27.9 & 15.3 & 14.9 & 16.2 & 14.0 & 24.9 \\
        
		C-CONet~\cite{wang2023openoccupancy} & C & 20.1 & 12.8 & 13.2 & 8.1 & 15.4 & 17.2 & 6.3 & 11.2 & 10.0 & 8.3 & 4.7 & 12.1 & 31.4 & 18.8 & 18.7 & 16.3 & 4.8 & 8.2  \\
        
		L-CONet~\cite{wang2023openoccupancy} & L & 30.9 & 15.8 & 17.5 & 5.2 & 13.3 & 18.1 & 7.8 & 5.4 & 9.6 & 5.6 & 13.2 & 13.6 & 34.9 & 21.5 & 22.4 & 21.7 & 19.2 & 23.5  \\
		
		M-CONet~\cite{wang2023openoccupancy} & \makecell{C\&L} & 29.5 & 20.1 & 23.3 & 13.3 & 21.2 & 24.3 & 15.3 & 15.9 & 18.0 & 13.3 & 15.3 & 20.7 & 33.2 & 21.0 & 22.5 & 21.5 & 19.6 & 23.2   \\

		PointOcc~\cite{pointocc} & L & 34.1 & 23.9 & 24.9 & 19.0 & 20.9 & 25.7 & 13.4 & 25.6 & 30.6 & 17.9 & 16.7 & 21.2 & 36.5 & 25.6 &25.7 & 24.9 & 24.8 & 29.0  \\ 

        \midrule

        \rowcolor{nbicycle} PVP (Ours) & L & \textbf{37.0} & 25.8 & 29.9 & 16.5 & 23.6 & 30.4 & 12.5 & 23.5 & 32.5 & 17.0 & 20.0 & 23.9 & \textbf{41.7} & 26.8 & 28.4 & 26.7 & 28.2 & \textbf{30.5}  \\ 
        
       \rowcolor{nbicycle} PVP (Ours) & C\&L & 36.3 & \textbf{28.0} & \textbf{32.2} & \textbf{24.5} & \textbf{26.7} & \textbf{31.8} & \textbf{16.4} & \textbf{29.9} & \textbf{34.9} & \textbf{21.9} & \textbf{21.5} & \textbf{26.5} & 40.3 & \textbf{27.5} & \textbf{28.6} & \textbf{26.9} & \textbf{28.5} & 30.4  \\ 
  
		\bottomrule
	\end{tabular}
	\label{tab_occ_val}
\end{table*}

\section{Experiments} 

We assess our method on the OpenOccupancy~\cite{wang2023openoccupancy} benchmark for 3D semantic occupancy prediction. Section \ref{sec_exp_setup} and \ref{imd} detail the datasets and implementation specifics. Following that, in Section \ref{sec_exp_open_occ}, we compare our method against other SOTA approaches. Lastly, to explore the impact of various components and designs within our method, we perform ablation studies in Section \ref{ablation}.

\subsection{Experimental setup}
\label{sec_exp_setup}
\subsubsection{OpenOccupancy dataset} The OpenOccupancy dataset~\cite{wang2023openoccupancy} is based on the data collected in nuScenes dataset~\cite{caesar2020nuscenes}. The dataset comprises 700 point cloud sequences for training and 150 sequences for validation. Each sequence is approximately 20 seconds long, captured with a LiDAR frequency of 20Hz.
The dataset includes 28,130 annotated frames for training and 6,019 for validation. Data collection utilized a 32-lane LiDAR and six surrounding cameras. 
For 3D occupancy prediction, the primary metrics include the Intersection over Union (IoU), serving as the geometric metric to identify a voxel as either occupied or empty (treating all occupied voxels as one category), and the mean IoU (mIoU) across all classes, acting as the semantic metric \cite{wang2023openoccupancy, liu2023toothsegnet}.
The evaluation range of OpenOccupancy is set as [-51.2m, -51.2m, -5m] to [51.2m, 51.2m, 3m] for $X$, $Y$ and $Z$ axis, and the voxel resolution is [0.2m, 0.2m, 0.2m], resulting in a output volume of size $512\times512\times40$ for occupancy prediction.
On the OpenOccupancy Dataset \cite{wang2023openoccupancy}, our model adopts a perception range of [0.3m, 73.0m] for the $\rho$ axis, [$-\pi$, $\pi$] for the $\phi$ axis, and [-5m, 3m] for the \textit{z} axis.

\subsubsection{Compared Baselines} 
We compared our PVP with PointOcc \cite{pointocc}
, M-CONet \cite{wang2023openoccupancy}, L-CONet \cite{wang2023openoccupancy}, C-CONet \cite{wang2023openoccupancy}, JS3C-Net \cite{yan2020sparse}
, LMSCNet \cite{roldão2020lmscnet}, AICNet \cite{li2020anisotropic}, 3DSketch \cite{chen20203d}, TPVFormer \cite{huang2023tri}, MonoScene \cite{cao2022monoscene}, as shown in \autoref{tab_occ_val}. 
It's noteworthy that MonoScene \cite{cao2022monoscene} and TPVFormer \cite{huang2023tri}, along with C-CONet \cite{wang2023openoccupancy}, primarily use camera images as input. In contrast, 3DSketch \cite{chen20203d} and AICNet \cite{li2020anisotropic} incorporate a multi-modal input approach, utilizing both camera images and depth images. LMSCNet \cite{roldão2020lmscnet}, JS3C-Net \cite{yan2020sparse}, L-CONet \cite{wang2023openoccupancy}, and PointOcc \cite{pointocc} rely on LiDAR point cloud data as their input source. 
PVP is available in two versions: one version uses LiDAR point cloud data as its input, while the other version of PVP, along with M-CONet \cite{wang2023openoccupancy}, takes both camera images and LiDAR point cloud data as inputs.
All baseline implementations were aligned with the benchmark \cite{wang2023openoccupancy} to ensure fairness in comparison.

\subsection{Implementation Details}
\label{imd}
\noindent\textbf{Model Architecture.} 
For a fair comparison, we follow the same pipeline, including the training scheme and head design, as CONet~\cite{wang2023openoccupancy} for our evaluations.
For the voxelization, we transform the original point cloud to a sparse volume with the size of $(H, W, D)=(1024, 1344, 80)$, where three dimensions indicate the $\rho$, $\phi$ and \textit{z}, respectively.
In the 3D LiDAR backbone, we utilize the same backbone architecture as CONet but replace the basic block with our proposed PD-Conv block for enhanced processing.
The final volume representation output by the entire pipeline has dimensions $(\mathcal{H}, \mathcal{W}, \mathcal{D}) = (128, 168, 10)$, representing an 8x down-sampling from the initial volume size. The task occupancy head in PVP is configured in accordance with the specifications provided in \cite{wang2023openoccupancy}.

\noindent\textbf{Optimization.} 
During training, we employ the Adam optimizer with a weight decay of 0.01. A cosine learning rate scheduler is adopted, featuring a maximum learning rate of 3e-4 and a linear warm-up period for the first 500 iterations. For occupancy prediction, we follow the approach of OpenOccupancy \cite{wang2023openoccupancy}, incorporating an affinity loss to further optimize both geometric and semantic metrics.
Our models are trained for 25 epochs with a batch size of 8 across 8 V100 GPUs. Given resource constraints, we report the results of ablation studies based on models that were trained for five epochs.


\subsection{3D Semantic Occupancy Prediction Results}
\label{sec_exp_open_occ}
We assess the effectiveness of our PVP on the OpenOccupancy~\cite{wang2023openoccupancy} benchmark. 
As demonstrated in the \autoref{tab_occ_val}, our PVP model achieves significant performance improvements compared to previous methods across all input modalities.
Compared with the Multimodal M-CONet~\cite{wang2023openoccupancy}, PVP improves $\mathbf{mIoU}$ and $\mathbf{IoU}$ by 
$\textbf{7.9\%}$ and $\textbf{6.8\%}$
, respectively, which is a giant leap on performance. 
Compared with other polar-based methods such as PointOcc~\cite{pointocc}, which is a LiDAR-only method, the LiDAR-only PVP still outperforms PointOcc \cite{pointocc} on both $\mathbf{mIoU}$ and $\mathbf{IoU}$ by 
$\textbf{1.9\%}$ and $\textbf{2.9\%}$
, respectively, which is still a considerable gain on performance.

        



\begin{table}[t]
\centering
\renewcommand\arraystretch{1.}
\caption{\textbf{Effects of different components in PVP.}    We show the 5 epochs results of mIoU and IoU on the nuScenes validation set.}
\label{tab_abl_comp}
\setlength{\tabcolsep}{10pt}
\begin{tabular}{ccc|c} 
\toprule
Polar & GRP & PD-Conv & IoU/mIoU(\%)                 \\ 
\hline\hline
      &     &         & 26.4/19.4                    \\
   \checkmark   &     &         & 26.0/20.1                    \\
   \checkmark   &  \checkmark   &         & 31.8/22.0                    \\
    \checkmark  &     &    \checkmark     & 33.2/23.8                    \\
    \checkmark  &  \checkmark   &    \checkmark     & \textbf{35.2}/\textbf{25.1}  \\
\bottomrule
\end{tabular}
\end{table}

\begin{table}[h]

 \caption{\textbf{Comparison of different attention mechanisms in the propagation.} We show the 5 epochs results on the nuScenes validation set. Stuff mIoU is the mIoU of following classes: \textcolor{ndriveable}{$\blacksquare$}
		\textcolor{nother}{$\blacksquare$}
		\textcolor{nsidewalk}{$\blacksquare$}
		\textcolor{nterrain}{$\blacksquare$}
		\textcolor{nmanmade}{$\blacksquare$}
		\textcolor{nvegetation}{$\blacksquare$}.}
 \vspace{-3mm}
\renewcommand\arraystretch{1.1}
\begin{center}
 \setlength{\tabcolsep}{10pt}
 \resizebox{\linewidth}{!}{
    \begin{tabular}{c|c|c|c|c}
    \toprule
    {Coordinate} & {Attention} & {IoU(\%)} & {mIoU(\%)} & stuff mIoU(\%)\\
    \hline
    \hline
    Cartesian & - & 26.4  & 19.4 & 21.8  \\
    Polar & - & 26.0  & 20.1 & 21.0  \\
    Polar & Channel~\cite{fan} & 26.5 & 20.2 & 21.5 \\
    Polar & Swin~\cite{swin} & 26.1 & 20.2 & 21.1 \\
    Polar & GRR~\cite{nie2023partner} & 26.1  & 20.2 & 21.1 \\
    Cartesian & GRP & 26.8  & 19.6 & 22.3 \\
    Polar & GRP & \textbf{31.8}  & \textbf{22.0} & \textbf{25.8} \\
    \bottomrule
    \end{tabular}
    }
    \end{center}
    \vspace{-4mm}
    \label{tab_grp}
\end{table}

\begin{table}[t]
\renewcommand\arraystretch{1.1}
	\caption{\textbf{Efficiency analysis of PVP.} GPU Mem. represents the GPU memory consumption at the training stage. OpenOccupancy denotes the baseline method in \cite{wang2023openoccupancy}.}
		\begin{center}
            \setlength{\tabcolsep}{10pt}
            \resizebox{\linewidth}{!}{
		\begin{tabular}[b]{c|c|c|c|c}
			\toprule
			 Method & Modal & GPU Mem. & GFLOPs & IoU/mIoU(\%)
			\\
            \midrule
			\midrule
             OpenOccupancy \cite{wang2023openoccupancy} & L & 22 GB & 5899  & 30.7/15.0 \\
             OpenOccupancy \cite{wang2023openoccupancy} & C\&L & 40 GB & 13117 & 29.3/19.8 \\
             CONet \cite{wang2023openoccupancy} & L & 8.5 GB & 810 & 30.9/15.8 \\
             CONet \cite{wang2023openoccupancy} & C\&L & 24 GB & 3066 & 29.5/20.1 \\
             \hline
             PVP & L & 7.9 GB & 801 & \textbf{37.0}/25.8 \\
             PVP & C\&L & 23.5 GB & 2950 & 36.3/\textbf{28.0} \\
			\bottomrule
		\end{tabular}
  }
      \end{center}
	\label{tab_efficiency}
\end{table}

\begin{table}[t]

 \caption{\textbf{Comparison of different structures of plane decompose convolution.} We show the 5 epochs results on the nuScenes val. set. Stuff mIoU is the mIoU of following classes: \textcolor{ndriveable}{$\blacksquare$}
		\textcolor{nother}{$\blacksquare$}
		\textcolor{nsidewalk}{$\blacksquare$}
		\textcolor{nterrain}{$\blacksquare$}
		\textcolor{nmanmade}{$\blacksquare$}
		\textcolor{nvegetation}{$\blacksquare$}. extra Mem. represents the extra GPU memory consumption caused by pd-convs at the training stage.}
 \vspace{-5mm}
\renewcommand\arraystretch{1.1}
\begin{center}
    \setlength{\tabcolsep}{10pt}
    \resizebox{\linewidth}{!}{
    \begin{tabular}{c|c|c|c|c}
    \shline
    {Model} & extra Mem. & {IoU(\%)} & {mIoU(\%)} & stuff mIoU(\%)\\
    \hline
    \hline
    Naive & - & 26.0 & 20.1 & 21.1  \\
    Assym~\cite{zhu2021cylindrical} & - 0.1 GB & 32.4  & 22.7 & 26.8  \\
    \hline
    (a) & + 0.3 GB  & 33.2  & 23.1 & 27.3  \\
    (b) & + 0.4 GB  & 33.2 & 23.3 & 27.4 \\
    (c) & + 0.3 GB  & 33.2 & 23.2 & 27.3 \\
    (d) & + 0.5 GB  & 33.2 & 23.3 & 27.4 \\
    \shline
    \end{tabular}
    }
    \end{center}
    \vspace{-2em}
    \label{tab_pd_design}
\end{table}

\subsection{Ablation studies}
\label{ablation}
\noindent\textbf{Effectiveness of different components in PVP.}
In \autoref{tab_abl_comp}, we investigate the effectiveness of each component in our proposed method.
The data indicates that directly transitioning to polar representation yields only a 0.7\% increase in mIoU, while it actually results in a 0.4\% decrease in IoU.
The GRP module, when applied independently to the predictor, enhances performance by 1.9\% in mIoU and 5.8\% in IoU compared to the raw baseline.
Additionally, adopting the PD-Conv module leads to a significant performance improvement, achieving a 3.7\% increase in mIoU and a 7.2\% increase in IoU compared to the raw baseline.
Combining the two key designs, we can obtain a performance gain of $9.2\%$ IoU and $5.0\%$ mIoU compared to the raw baseline.


\noindent\textbf{Effectiveness of GRP.}
In \autoref{tab_grp}, {we assess various attention mechanisms employed for feature propagation.
An attempt to enlarge the receptive field is to introduce channel-wise attention~\cite{fan} during polar feature recalibration, by which distorted representations can also be integrated globally.
Unfortunately, this practice attains limited improvement with a $0.1\%$ gain on mIoU.}
A plausible explanation for the limited success of this approach is that aligning features at the global level, while sacrificing substantial structural information, tends to introduce more background noise and results in lower-quality feature representations.
Another attempt to enlarge the receptive field efficiently is to introduce GRR Module from PARTNER~\cite{nie2023partner}.
However, experimental outcomes reveal that this method yields minimal improvement, with only a 0.1\% gain in mIoU. This lack of effectiveness may be attributed to an insufficient number of representative voxels.
Contrary to 3D detection, 3D occupancy prediction struggles with long-range feature propagation needed to identify road structures. Given its focus on foreground objects, the representation voxel mechanism may fall short in detecting ``stuff'' classes, such as roads and buildings, which require a broader spatial understanding.
Additionally, we experimented with utilizing Swin attention~\cite{swin} to facilitate feature propagation within local regions. However, the marginal performance improvement observed in the ablations (a $0.1\%$ gain in mIoU) is constrained by background noise and a restricted receptive field.

\noindent\textbf{Efficiency of PVP.}
To validate the efficiency of our PVP, we conducted an ablation study to compare it with SOTA methods. As shown in \autoref{tab_efficiency}, relative to CONet \cite{wang2023openoccupancy} with multi-modal input, it is clear that PVP not only achieves a significant 7.9\% increase in mIoU but also reduces memory usage by 0.5GB and marginally lowers computational costs.

\noindent\textbf{Effectiveness of PD-Conv.}
To assess the efficacy of the PD-Conv mechanism, we undertake ablation studies to identify the sources of improvement. As illustrated in ~\autoref{tab_pd_design}, we evaluate various configurations of PD-Conv (serial (a), parallel (b), or a combination of both (c) and (d)). It is observed that differences in the structure of the decomposed 2D convolution result in only marginal performance variations. This consistent performance suggests that the primary gain in effectiveness stems from the simplified patterns on each plane due to decomposition. The three 2D convolutions ensure the stability of performance.

\section{Conclusions}

In this study, we revisit the challenges of feature distortion and feature propagation in the polar representation of 3D space, which significantly hampers the effectiveness of polar-based 3D semantic occupancy predictors. To address these issues, we propose PVP, a novel 3D occupancy predictor designed for polar coordinates. PVP mitigates feature distortion through the PD-Conv and enhances feature propagation by implementing a long-range attention mechanism with the proposed GRP module. Our results demonstrate that the proposed method substantially outperforms previous polar-based efforts and achieves competitive outcomes when compared to SOTA methods in 3D occupancy prediction. We hope our work will inspire further exploration into polar-based 3D perception technologies.

\newpage

{\small

\bibliographystyle{unsrt}
\bibliography{main}
}

\end{document}